# REPRESENTATION LEARNING ON EVENT STREAM VIA AN ELASTIC NET-INCORPORATED TENSOR NETWORK

*Beibei Yang[1], Weiling Li[1], Yan Fang[2]*

[1]School of Computer Science and Technology, Dongguan University of Technology, Dongguan, China
[2]Department of Electrical and Computer Engineering, Kennesaw State University, Marietta, GA, USA

## ABSTRACT

Event cameras are neuromorphic sensors that capture asynchronous and sparse event stream when per-pixel brightness changes. The state-of-the-art processing methods for event signals typically aggregate events into a frame or a grid. However, events are dense in time, these works are limited to local information of events due to the stacking. In this paper, we present a novel spatiotemporal representation learning method which can capture the global correlations of all events in the event stream simultaneously by tensor decomposition. In addition, with the events are sparse in space, we propose an Elastic Net-incorporated tensor network (ENTN) model to obtain more spatial and temporal details about event stream. Empirically, the results indicate that our method can represent the spatiotemporal correlation of events with high quality, and can achieve effective results in applications like filtering noise compared with the state-of-the-art methods.

## 1. INTRODUCTION

Event cameras, also known as Dynamic Vision Sensors (DVS) [1], are commercially available after years of development. They detect changes in brightness and generate events when the brightness change exceeds a threshold. These events contain spatiotemporal information about the observed objects. Event cameras have high temporal resolution at the microsecond level, enabling vision in poor lighting and quick responses. They perform well in tasks like high-speed motion estimation and high dynamic range mapping.

To fully comprehend the events, researchers have started employing representation learning-based methods to extract more informative content from the event stream. The mainstream processing methods for event data can be categorized into event-by-event methods and group-of-events methods. Aggregating events into frames based on specific time periods and analyzing the event stream frame by frame is a commonly employed technique [2-5] for representing the event stream. This approach facilitates a more comprehensive understanding of the dynamic changes captured by event cameras. However, compared to analyzing events frame by frame, it is important to note that the event stream exhibits a dense temporal nature. Therefore, employing a representation learning model that can capture the global correlations among all events in the event stream would provide a more comprehensive understanding of the underlying data.

It is worth noting that the event stream can be represented as a 3rd-order tensor. Tensor decomposition models have demonstrated their effectiveness in capturing the correlations among tensor elements [6-9]. Furthermore, the spatiotemporal correlation of events is closely related to the motion patterns of the corresponding objects or targets [10-12]. However, an event tensor only contains pixels which the brightness change exceeds the set threshold, neglecting certain parts of the moving targets, as illustrated in Fig. 1. Hence, using tensor completion as the task objective for tensor decomposition can better utilize observed events.

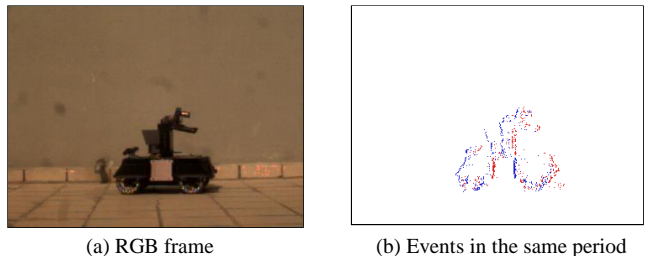

(a) RGB frame          (b) Events in the same period
**Fig. 1.** RGB frame and its corresponding events from VisEvent [11].

Based on the aforementioned findings, we introduce a novel model Elastic Net-incorporated tensor network (ENTN) which combines the concepts of Elastic-Net regularization and tensor network techniques for performing spatiotemporal representation learning on event streams with twofold ideas:
a) We propose a fully-connected 3rd-order tensor network that treats tensor completion as its objective, enabling spatiotemporal representation learning on event streams.
b) We integrate Elastic-net regularization into the process of updating latent factors to improve feature selection, considering the sparsity of events in space.

Experimental evaluations conducted both on public and self-collected event datasets demonstrate that the proposed model has a strong ability to represent the spatiotemporal correlation of events and can achieve effective results in applications like filtering noise.

## 2. PRELIMINARIES

**Notation.** In this paper, we denote scalars, vectors, metrics and 3rd-order tensors by $x$, **x**, **X**, $X$, respectively. $\mathbf{x}(\alpha)$, $\mathbf{X}(\alpha, \beta)$ and $X(\alpha, \beta, \gamma)$ are used to denote elements in **x**, **X** and $X$.

**Event tensor.** DVS outputs event stream when it is triggered by brightness change over the threshold. Each event in the stream contains a $(i, j, t)$ 3-tuple, where $t$ is the timestamp in microseconds, $i$ and $j$ are the pixel address. To represent spatiotemporal feature of sparse event, in this study, an event stream is firstly converted into a 3rd-order tensor, as shown in Fig. 2. Considering a DVS monitors a field of view of $I \times J$ for a time of $t$, its output stream can be transformed into a 3rd-order tensor, e.g., $E \in R^{I \times J \times N}$, by dividing $t$ evenly into $N$ segments. E's element, e.g., $e_{ijn}$, is set to 1 if pixel in $(i, j)$ has been triggered within the $n$-th time period.

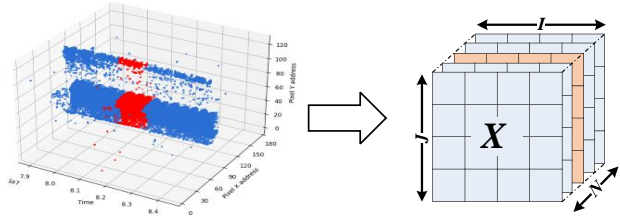

**Fig. 2.** Event stream representation by tensor.

**Fully-connected 3rd-order tensor network (F3TN).** It is a special case of fully-connected tensor network (FCTN) for tensor decomposition. It decomposes a 3rd-order tensor $E^{I \times J \times T}$ into three small-sized 3rd-order factors, i.e., $G_i \in R^{I \times f \times f}$, $G_j \in R^{f \times J \times f}$ and $G_n \in R^{f \times f \times N}$, and each factor needs to conduct tensor contraction with all other factors. Mathematically, the element-wise form of the F3TN decomposition can be formulated as

$$\hat{E} = F3TN(G_i, G_j, G_n)$$
$$\Rightarrow \hat{e}_{ijn} = \sum_{x=1}^{f}\sum_{y=1}^{f}\sum_{z=1}^{f}\{G_i(i,x,y)G_j(x,j,z)G_n(y,z,n)\}, \quad (1)$$

where $f$ is the latent factor dimension, $\hat{e}_{ijk}$ in $\hat{E}$ is the approximate of $e_{ijk}$ in $E$. A tensor decomposition model is a minimization of the Euclidean distance [13-15] between $E$ and $\hat{E}$ by estimating $G_i$, $G_j$ and $G_n$ as follows:

$$\arg\min_{G_i, G_j, G_n} \frac{1}{2}\|E - \hat{E}\|_F^2. \quad (2)$$

## 3. METHOD

In this section, we provide the details of the Elastic Net-incorporated fully-connected tensor network (ENTN) model for event stream representation. ENTN model takes event tensor, i.e., $E$, as its input and searches for latent factor tensors, i.e., $G_i$, $G_j$ and $G_n$ for solving following objective function:

$$\arg\min_{X, G_i, G_j, G_n} \frac{1}{2}\|X - F3TN(G_i, G_j, G_n)\|_F^2. \quad (3)$$

where $X$ is the target tensor based on the assumption that there are pixels whose brightness changes do not exceed the threshold, but they have spatiotemporal correlation with known events. Note that all optimization variables are coupled with each other, the framework of PAM [16] is utilized to solve (3). With it, the solution can be obtained by alternately updating.

Besides, recall that the known events are sparse in space, for better representation accuracy, we incorporate elastic-net regularization [17-20] into F3TN. Therefore, the optimization problem in (3) can be optimized by alternately updating the following set of equations:

$$\begin{cases} G_i^{(s+1)} = \arg\min_{G_i} \left\{ \frac{1}{2} \left\| X^{(s)} - F3TN\left(G_i, G_j^{(s)}, G_n^{(s)}\right) \right\|_F^2 \right. \\ \left. + \frac{\lambda_2}{2} \left\| G_i - G_i^{(s)} \right\|_F^2 + \lambda_1 \left| G_i - G_i^{(s)} \right|_F \right\}, \\ G_j^{(s+1)} = \arg\min_{G_j} \left\{ \frac{1}{2} \left\| X^{(s)} - F3TN\left(G_i^{(s+1)}, G_j, G_n^{(s)}\right) \right\|_F^2 \right. \\ \left. + \frac{\lambda_2}{2} \left\| G_j - G_j^{(s)} \right\|_F^2 + \lambda_1 \left| G_j - G_j^{(s)} \right|_F \right\}, \\ G_n^{(s+1)} = \arg\min_{G_n} \left\{ \frac{1}{2} \left\| X^{(s)} - F3TN\left(G_i^{(s+1)}, G_j^{(s+1)}, G_n\right) \right\|_F^2 \right. \\ \left. + \frac{\lambda_2}{2} \left\| G_n - G_n^{(s)} \right\|_F^2 + \lambda_1 \left| G_n - G_n^{(s)} \right|_F \right\}, \\ X^{(s+1)} = \arg\min_X \left\{ \frac{\lambda_2}{2} \left\| X - X^{(s)} \right\|_F^2 \right. \\ \left. + \frac{1}{2} \left\| X - F3TN\left(G_i^{(s+1)}, G_j^{(s+1)}, G_n^{(s+1)}\right) \right\|_F^2 \right\}, \end{cases} \quad (4)$$

where $\lambda_1$ and $\lambda_2$ are coefficients of L1 and L2 regularization, respectively.

For the sub-problems, it is difficult to calculate the derivative of their objective functions in the tensor form. According to [21], we firstly rewrite $G_i$ as the following matrix form as follows:

$$\arg\min_{G_i} \left\{ \frac{1}{2} \left\| \mathbf{X}^{(s)} - \mathbf{G}_i \mathbf{H}_i^{(s)} \right\|_F^2 + \frac{\lambda_2}{2} \left\| \mathbf{G}_i - \mathbf{G}_i^{(s)} \right\|_F^2 + \lambda_1 \left| \mathbf{G}_i - \mathbf{G}_i^{(s)} \right|_F \right\}, \quad (5)$$

where $\mathbf{H}_i = \Gamma(H_i)$ for $\Gamma()$ is the Genunfold operator proposed in [22] and $H_i^{(s)} = F3TN(G_j^{(s)}, G_n^{(s)})$. Note that (5) is differentiable, and thus its solution can be obtained by solving the following Sylvester matrix equation:

$$\mathbf{G}_i \mathbf{H}_i^{(s)} \mathbf{H}_i^{(s)} + \lambda_2 \mathbf{G}_i = \mathbf{X}^{(s)} \mathbf{H}_i^{(s)} + \lambda_2 \mathbf{G}_i^{(s)} + \lambda_1 \mathbf{1} \quad (6)$$

where $\mathbf{1}$ is a matrix where $\mathbf{1}(\alpha, \beta)=1$ for $\alpha=\beta$ and $\mathbf{1}(\alpha, \beta)=0$ for $\alpha\neq\beta$. Therefore, we have the update rules for $G_i$, $G_j$ and $G_n$ as follows:

$$\begin{cases} \mathbf{G}_i^{(s+1)} = \dfrac{\mathbf{X}^{(s)} \mathbf{H}_i^{(s)} + \lambda_2 \mathbf{G}_i^{(s)} + \lambda_1 \mathbf{1}}{\mathbf{H}_i^{(s)} \mathbf{H}_i^{(s)} + \lambda_2 \mathbf{I}}, \\ \mathbf{G}_j^{(s+1)} = \dfrac{\mathbf{X}^{(s)} \mathbf{H}_j^{(s)} + \lambda_2 \mathbf{G}_j^{(s)} + \lambda_1 \mathbf{1}}{\mathbf{H}_j^{(s)} \mathbf{H}_j^{(s)} + \lambda_2 \mathbf{I}}, \\ \mathbf{G}_n^{(s+1)} = \dfrac{\mathbf{X}^{(s)} \mathbf{H}_n^{(s)} + \lambda_2 \mathbf{G}_n^{(s)} + \lambda_1 \mathbf{1}}{\mathbf{H}_n^{(s)} \mathbf{H}_n^{(s)} + \lambda_2 \mathbf{I}}, \end{cases} \quad (7)$$

Moreover, by implementing the same process, we have the update rule for $X$ as follows:

$$X^{(s+1)} = \frac{F3TN\left(G_i^{(s+1)}, G_j^{(s+1)}, G_n^{(s+1)}\right) + \lambda_2 X^{(s)}}{1 + \lambda_2}. \quad (8)$$

The whole process of representing event stream with proposed method is summarized in Algorithm 1.

**Algorithm 1: Event stream representation with ENTN**

---
**Input**: The 3rd-order tensor $E \in R^{I \times J \times N}$, the maximal F3TN-rank $f^{max}$.
**Initialization**: $s=0$, $s^{max}=1000$, $X^{(0)}=E$, the initial F3TN-rank $f=$ max$\{1, f^{max}-5\}$, $G_i$, $G_j$ and $G_n$
**Output**: The latent factors $G_i$, $G_j$ and $G_n$.

1: **while** not converged and $s < s^{max}$ **do**
2:    Update $G_i$, $G_j$ and $G_n$. via (7).
3:    Update $X$ via (8).
4:    Let $f=\min\{f+1, f^{max}\}$ and expand $G_i$, $G_j$ and $G_n$. if
$$\left\| X^{(s+1)} - X^{(s)} \right\|_E / \left\| X^{(s)} \right\|_E < 10^{-2}.$$
5:    Check the convergence condition:
$$\left\| X^{(s+1)} - X^{(s)} \right\|_E / \left\| X^{(s)} \right\|_E < 10^{-3}.$$
6:    $s=s+1$.
7: **end while**

## 4. EXPERIMENTS AND RESULTS

In this section, we try to understand the performance of proposed ENTN by answering the following questions experimentally:
   **Q1**: Whether event stream can be represented by tensor decomposition?
   **Q2**: Can Elastic-net regularization improve the performance of tensor network for representing event stream?

### 4.1. Experiments Settings

**Implementation details.** To verify the correctness of representation results, we employ a classifier, e.g., Support Vector Machine (SVM) [23], to classify different kind of events. In our experiments, a part of events, which have the same trail characteristics, are tagged with same label. For instance, for an event, we vectorize its corresponding latent factors obtained by tensor decomposition [24-26] and input it into SVM for training and testing.

Among the tested event tensors, the first 60% of the time frames are used to train the classifier and the remaining 40% is used for testing. Therefore, classification accuracy can indirectly reflect the spatiotemporal representation ability of the model. For all representation models based on event tensor decomposition, we use the same process for testing.

**Data sets.** Four datasets are employed in our experiments, whose details are summarized in Table 1. D1 and D2 are collected in our laboratory with iniVation DAVIS 346. D1 and D2 both contain different motion trajectories of two objects, and we have labeled them differently. We randomly generate background noise for them while labeling the events of the moving object differently from the noise events.

**Table 1.** Details of involved datasets

| Data | Coordinate Scale | # of frames | # of events | # of density |
|---|---|---|---|---|
| D1 | 346×260 | 100 | 56205 | 0.62% |
| D2 | 346×260 | 100 | 56205 | 0.62% |

**Evaluation Metrics.** As state in Methodology, we measure the performance of involved models by classification accuracy. Hence, we adopt Area Under the Curve (AUC) as the evaluation metrics as did in [25].

**Involved models.** Table 2 lists the models involved in the experiments. M2-M4 are the most representative methods for tensor representation. All experiments are implemented on a platform with a CORE-i5 2.50 GHz CPU and RTX 2050 GPU. M1-M4 are implemented with MATLAB.

**Table 2.** Involved models in our experiments

| No. | Description |
|---|---|
| M1 | ETFN, a model proposed in this work. |
| M2 | FCTN, a fully-connected tensor network for tensor completion proposed in [21]. |
| M3 | TR, a tensor ring-based model for tensor completion proposed in [26]. |
| M4 | CP, a Canonical Polyadic-based model for tensor completion proposed in [27]. |

**Implementation settings.** All experiments are implemented on a platform with a CORE-i5 2.50 GHz CPU and RTX 2050 GPU. M1-M4 are implemented with MATLAB.

## 4.2. Comparison Results

Experimental results are listed in Table 3 and 4. From the experimental results, we have the answers for Q1-Q2:

**Answer to Q1 and Q2: Tensor decomposition-based models represent event stream well and with Elastic-net regularization incorporation, the performance of the tensor network can be improved.** As show in Table 3, M1 to M4, which are tensor decomposition-based models, achieve high AUC on D1 and D2, which consist of two moving object with different trails. For example, on D1, M1 to M4 achieve AUC of 92.56%, 88.44%, 86.78% and 62.41%, respectively. From these results, we conclude that with the features obtained by tensor decomposition-based models, a classifier can effectively distinguish two objects with different trails.

**Table 3.** Representation results achieved by involved TD-based models

| Data | Models | AUC(%)↑ |
|---|---|---|
| D1 | M1 | 92.36⊙ |
|    | M2 | 88.44○ |
|    | M3 | 86.78 |
|    | M4 | 62.41 |
| D2 | M1 | 92.68⊙ |
|    | M2 | 86.48 |
|    | M3 | 89.53○ |
|    | M4 | 60.98 |

⊙ denotes the best results and ○ the second best.

Besides, with Elastic-net regularization, M1's performance can be improved. To better understand the effect of regularization incorporation, parameter sensitive tests have been taken. As shown in Fig. 3, the AUC achieved by M1 is variable with different regularization parameters. The AUC gap (i.e., $100\% \times (AUC_{highest} - AUC_{lowest})/AUC_{highest}$) is 12.66% and 33.3%, between $\lambda_1=0.6$ and $\lambda_2=0.6$ on D1, respectively. M1 can achieve better results than M2. For instance, as shown in table 3, on D1, M1's AUC is 92.36%, 3.92% higher than AUC at 88.44% by M2. On D2, compared with AUC achieved by M2, M1's improvement is 6.2%.

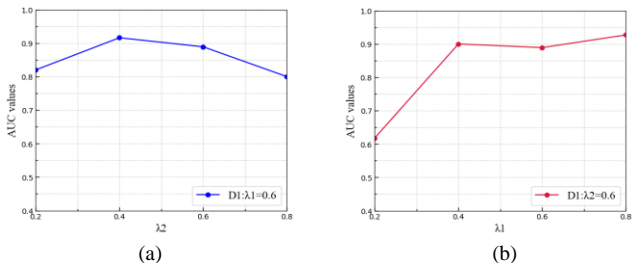

(a)      (b)
**Fig. 3.** AUC values with varying $\lambda_1$ and $\lambda_2$ of ENTN

## 5. CONCLUSION

In this work, we propose an Elastic Net-incorporated tensor network model that combines Elastic-net regularization with a fully-connected 3rd-order tensor network. The proposed mode demonstrates outstanding performance in representing the spatiotemporal correlation of events.